\documentclass[conference]{IEEEtran}
\IEEEoverridecommandlockouts
\usepackage{cite}
\usepackage{amsmath,amssymb,amsfonts}
\usepackage{algorithmic}
\usepackage{graphicx}
\usepackage{textcomp}
\usepackage{hyperref}
\usepackage{booktabs}
\usepackage{bbding}
\usepackage{pifont}

\usepackage[utf8]{inputenc} 
\usepackage[T1]{fontenc}    
\usepackage{url}            
\usepackage{nicefrac}       
\usepackage{microtype}      
\usepackage[normalem]{ulem}
\usepackage{multirow}
\usepackage{makecell}
\usepackage{stfloats}
\usepackage{wrapfig}
\usepackage{enumitem}
\usepackage{scalerel}
\usepackage{color}
\usepackage[dvipsnames]{xcolor}

\usepackage{algorithm}

\DeclareMathOperator*{\argmax}{argmax}

\def\BibTeX{{\rm B\kern-.05em{\sc i\kern-.025em b}\kern-.08em
    T\kern-.1667em\lower.7ex\hbox{E}\kern-.125emX}}
\begin{document}

\title{Render-and-Compare: Cross-view 6-DoF Localization from Noisy Prior\\
}


\author{\IEEEauthorblockN{ Shen Yan\IEEEauthorrefmark{1}, Xiaoya Cheng\thanks{$*$ Equal contribution.}\IEEEauthorrefmark{1}, Yuxiang Liu, Juelin Zhu, Rouwan Wu, Yu Liu, Maojun Zhang\thanks{\IEEEauthorrefmark{2} Corresponding author.}\IEEEauthorrefmark{2}}
\IEEEauthorblockA{\textit{College of Systems Engineering, National University of Defense Technology, ChangSha, China}
 \\
\{yanshen12, chengxy, liuyuxiang17, zhujuelin, wurouwan97, jasonyuliu, mjzhang\}@nudt.edu.cn}
}

\maketitle


%
\begin{abstract}
Despite the significant progress in 6-DoF visual localization, researchers are mostly driven by ground-level benchmarks. Compared with aerial oblique photography capture, ground-level map collection lacks scalability and complete coverage. In this work, we propose to go beyond the traditional ground-level setting and exploit cross-view 6-DoF localization from aerial to ground.
We address this problem by formulating camera pose estimation as an iterative render-and-compare pipeline and enhancing the algorithm robustness through augmenting seeds from noisy initial priors. As no public dataset exists for the studied problem, we have collected a new dataset that provides a variety of cross-view images from smartphones and low-altitude drones and developed a semi-automatic system to acquire ground-truth poses for query images. 
We benchmark our method as well as several state-of-the-art baselines and demonstrate that our method outperforms other approaches by a large margin. Code is available at \href{https://github.com/Choyaa/Render2Loc}{https://github.com/Choyaa/Render2Loc}.
\end{abstract}

\section{Introduction}
\label{sec:intro}

Visual localization aims to compute the camera pose for a given image relative to a known scene. Solving the problem is vital to many important applications, such as cars self-driving, UAV navigation, and augmented and virtual reality systems.

The majority of existing image localization methods~\cite{sarlin2019coarse, detone2018superpoint, sarlin2020superglue, sun2021loftr, zhou2021patch2pix, panek2022meshloc} infer camera location and orientation using feature matching~\cite{detone2018superpoint, sarlin2020superglue, sun2021loftr, zhou2021patch2pix} between the query image and a database of reference images from a similar perspective, typically ground-to-ground~\cite{sattler2018benchmarking, zhang2021reference}. The inherent limitation of these approaches is that building a complete and uniform reference map on the ground is complex and time-consuming.

Recent research~\cite{shi2019spatial, yang2021cross} has investigated the problem of cross-view geo-localization, which localizes ground-level query images by matching them against easily accessible aerial views. However, due to a lack of reference 3D models, such methods can only estimate a camera’s 3-DoF pose, namely, x-y coordinate position and azimuth angle. With the rapid development of 3D reconstruction techniques, it is now possible to build city-scale digital twins employing aerial oblique photography, such as Google Earth. In this paper, we thus explore the possibility of utilizing aerial 3D reconstruction to conduct cross-view 6-DoF localization.  

\begin{figure}[t]
   \centering
   \includegraphics[width=0.98\linewidth]{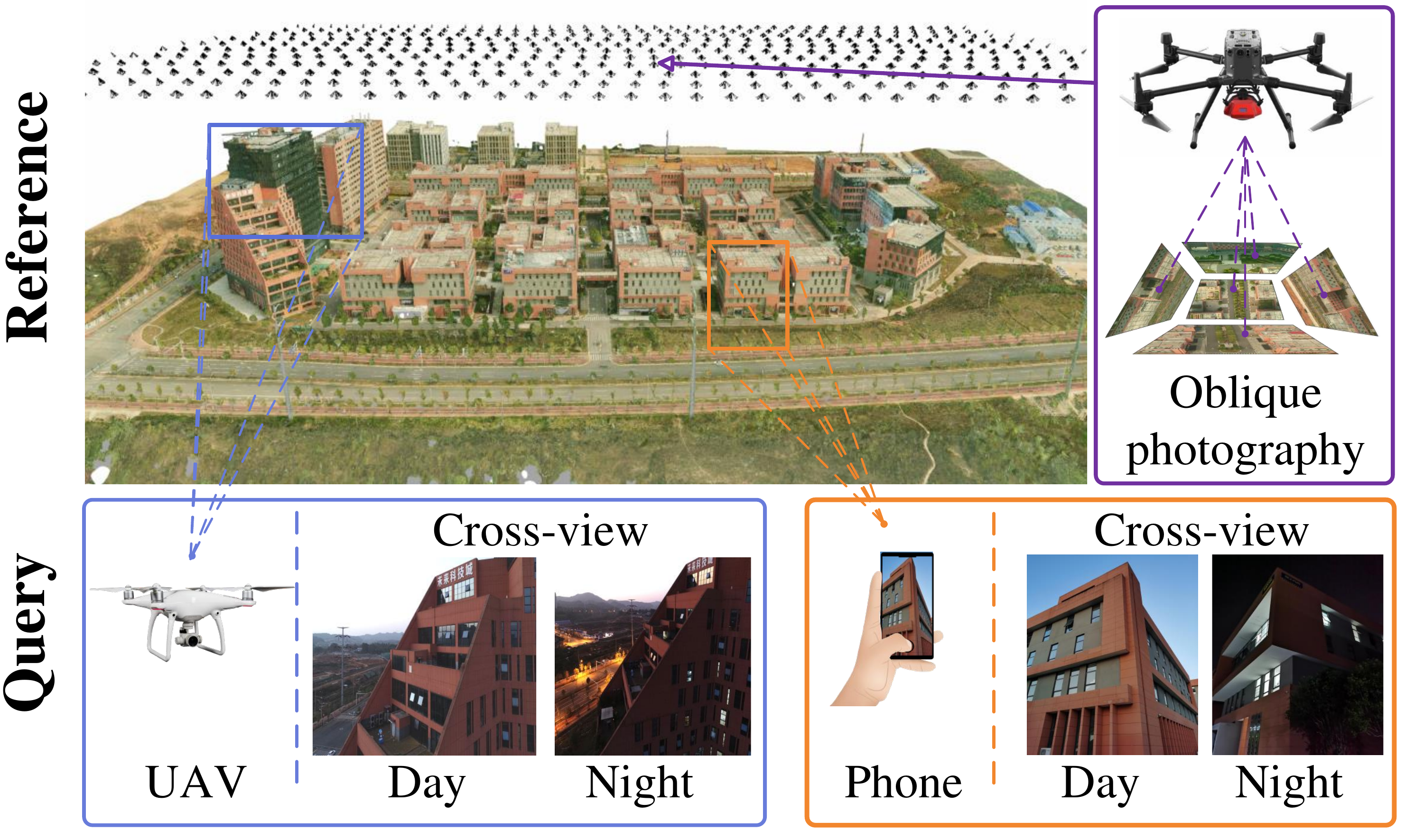}

   \caption{\textbf{Cross-view 6-DoF localization.} The proposed benchmark dataset \textit{AirLoc} exhibits drastic view changes between query and reference. The reference map is captured using a pentacular oblique camera above $100$ meters, while the query images are sampled close to the ground with small drones and smartphones, respectively. Query images include real Day-and-Night environments.}
   \label{fig:dataset}
\end{figure}   


This is quite an arduous task because, unlike ground-to-ground visual localization, image retrieval and feature matching between cross-view images are inherently difficulty as ground-to-aerial images capture totally distinctive appearances. To solve the problem, inspired by recent view synthesis work~\cite{panek2022meshloc}, we propose a novel localization framework \textit{Render-and-Compare}. Given an initial pose lead by the device’s sensor prior (i.e., GPS, compass, and gravity direction), our core idea is to move a virtual camera sequentially until the camera-observed scenarios align with the query image. Specifically, we first randomly sample several seeds around the sensor pose as candidates and produce synthetic images and depthmaps for all initial pose hypotheses based on textured mesh rendering. Then, we establish correspondences between the query image and synthetic candidates and select one with the largest inliers. According to pixel back-projecting in depthmap, 2D-2D correspondences can convert to 2D-3D matches, and the initial camera pose is recovered by a PnP RANSAC~\cite{chum2003locally, haralick1994review}. At last, the former pose estimation is iteratively updated by repeating render-and-matching operations and obtain a final result until the maximum iteration reaches.

To the best of our knowledge, there is no public dataset for 6-DoF localization under strong viewpoint changes. To facilitate the research of this area, we have collected a new benchmark dataset \textit{AirLoc}, as shown in Figure~\ref{fig:dataset}. Compared with other existing cross-view geo-localization datasets~\cite{zhai2017predicting, liu2019lending, zhu2021vigor}, \textit{AirLoc} owns the following advantages: 1) Textured 3D mesh references are built based on aerial oblique photography. 2) 6-DoF GT poses are assigned for query images instead of simply providing latitude, longitude, and azimuth angle. 3) The query image possesses different real conditions, including significant lighting variances (e.g., Day-and-Night). 4) The query image is captured by a commonly-used cellphone or drone, rather than panoramic imaging.  

We evaluate several image localization baselines and our method on the proposed dataset \textit{AirLoc}. The experiments demonstrate that our method outperforms state-of-the-art approaches by a large margin.

In summary, our main contributions include:
\begin{itemize}
    \item A novel \textit{Render-and-Compare} framework for robust and accurate visual localization under significant viewpoint changes.
    \item A new dataset for cellphone and UAV 6-DoF localization under great viewpoint conversions and strong illumination variations.
    \item Benchmarking existing methods and demonstrating the effectiveness of the proposed approach.
\end{itemize}

\section{Related Works}
\label{sec:related}


\subsection{Structured localization} 
State-of-the-art methods~\cite{sarlin2019coarse, detone2018superpoint, sarlin2020superglue, sun2021loftr, zhou2021patch2pix, panek2022meshloc} perform visual localization by establishing 2D-3D correspondences between the query images and sparse SfM points. The camera pose is recovered using a PnP solver~\cite{haralick1994review} inside a RANSAC loop~\cite{chum2003locally}. Besides, an intermediate image retrieval step~\cite{arandjelovic2016netvlad, yan2021image} is often applied to handle large-scale problems. However, the pipeline cannot guarantee the correctness of image matching in the case of cross-view localization. In this work, we show that the render-and-compare framework enables query and reference images to share similar geometric viewpoints, facilitating accurate feature matching.

\subsection{Cross-view Geo-localization} 
Recently, many methods~\cite{shi2019spatial, yang2021cross} treat the cross-view localization problem as a standard image retrieval task and solve it by aerial-to-ground image polar transformation and global feature embedding. However, these approaches output a 3-DoF camera pose at most, which is not sufficient to support VR/AR applications on mobile devices and precise control of drones. In contrast, our method can yield 6-DoF localization under significant viewpoint changes.

\subsection{Synthesis localization}
This work is not the first work to employ view synthesis for visual localization. The most related to our work are \cite{brejcha2020landscapear, panek2022meshloc, zhang2021reference}. Among them, \cite{brejcha2020landscapear} build a virtual panoramic map in mountainous terrain, and train local features to match images and this textured mesh. However, they show pretty coarse localization (hundreds of meters level), while we achieve centimeter-accurate localization. \cite{panek2022meshloc} show that meshes can be utilized to localize images from scratch and describe a full pipeline for this task. The drawback is that they suffer from terrible feature matching under notable viewpoint changes, while our render-and-compare framework ensures high-quality image matching by providing similar perspectives. \cite{zhang2021reference} render the scene from poses close to the ground-truth, and conduct iterative pose refinement for dataset GT pose labeling. We extend this idea to cross-view localization and successfully design a robust render-and-compare framework initialed from noisy sensor priors.

\subsection{Localization Datasets}
Most of existing 6-DoF localization benchmarks~\cite{sattler2018benchmarking, zhang2021reference} capture query and reference images from a similar perspective on the ground. The problem is that such data collection is time-consuming and cannot guarantee the integrity of the reference map. To this end, many cross-view geo-localization benchmarks are proposed recently, including CVUSA~\cite{zhai2017predicting}, CVACT~\cite{liu2019lending} and VIGOR~\cite{zhu2021vigor}, which aim to determine the locations of street-view query images by matching with GPS-tagged reference images from aerial view. Although these datasets can be easily scalable to the city level, they cannot solve the 6-DoF pose due to the lack of a 3D model. We combine the advantages of the two categories datasets mentioned above and introduce a new benchmark \textit{AirLoc}. \textit{AirLoc} builds a large-scale 3D model via aerial oblique photography, facilitating 6-DoF visual localization of any viewpoint within the bounds of this map. 

A detailed comparison of representative cross-view geo-localization datasets is given in Table~\ref{tab:dataset}.

\begin{table}[t]
    \centering
    \caption{\textbf{Overview of existing cross-view geo-localization datasets.} Several attributes are taken into account: GT pose (orientation, position and 6-DoF), seamless coverage on area of interest, and day-and-night conditions.}
    \resizebox{0.48\textwidth}{!}{
    \setlength\tabcolsep{2pt} 
    	\begin{tabular}{l||cccc} 
    		\Xhline{3\arrayrulewidth}
        	  & CVUSA~\cite{zhai2017predicting} & CVACT~\cite{liu2019lending} & VIGOR~\cite{zhu2021vigor} & \textit{AirLoc}  \\ 
    		\hline
    			GT orientation & \ding{51} & \ding{51}  & \ding{51} & \ding{51}    
    			\\
    			GT position & \ding{55} & \ding{51} & \ding{51} & \ding{51}                                               \\ 
    			GT 6-DoF pose & \ding{55} & \ding{55} & \ding{55} & \ding{51} 
                \\                                        
    			Seamless coverage & \ding{55} & \ding{55} & \ding{51} & \ding{51}                                           \\
    			Day-and-Night & \ding{55} & \ding{55} & \ding{55} & \ding{51}                                                            \\                     
    		\Xhline{3\arrayrulewidth}
    	\end{tabular}
    }
    \label{tab:dataset}
\end{table}

\section{Method}
\label{sec:method}
\begin{figure}[t]
   \centering
   \includegraphics[width=0.95\linewidth]{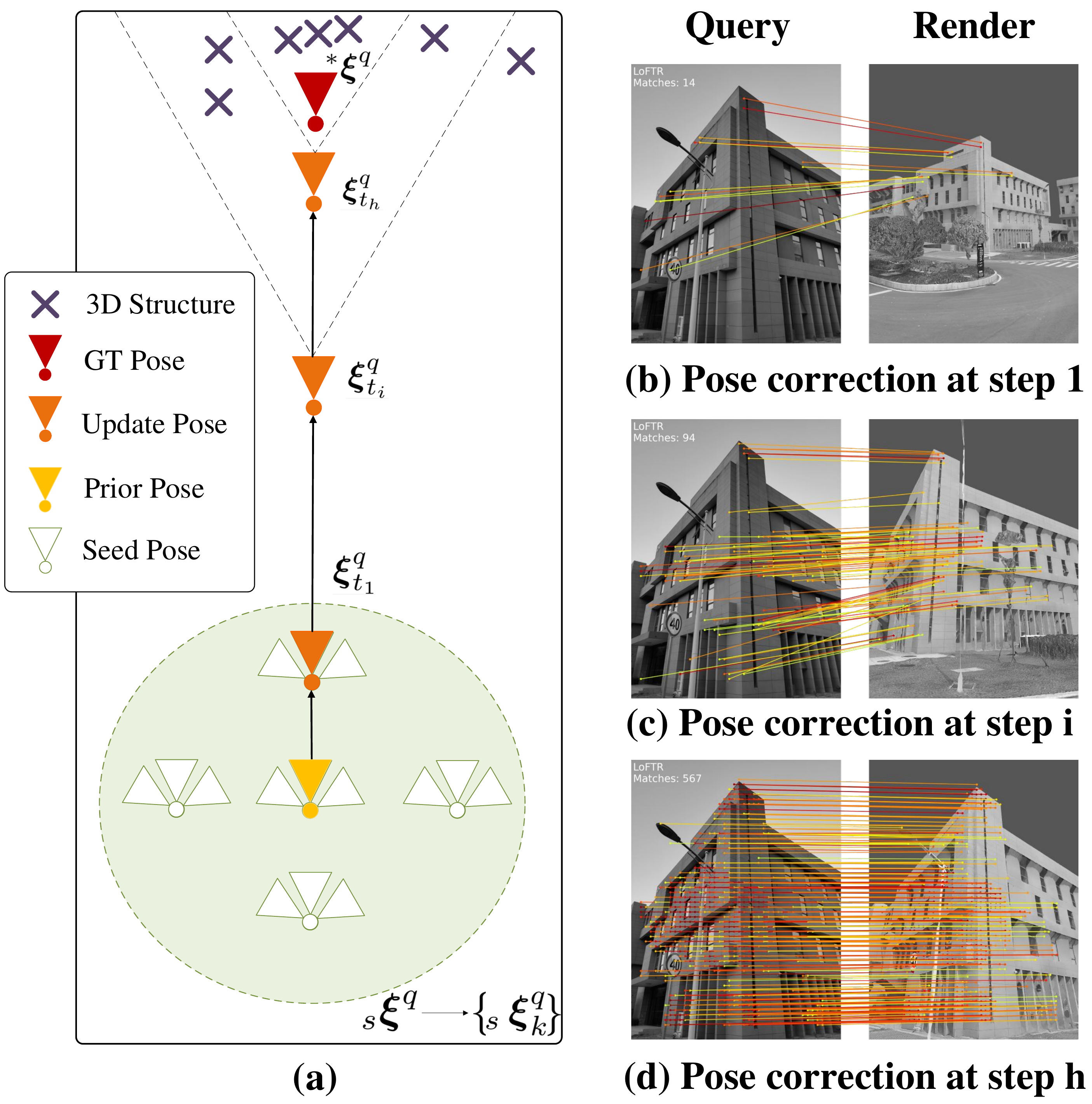}
   \caption{\textbf{Overview of the proposed method.} (a). For each prior pose $_s\boldsymbol{\mathcal{\xi}}^q$, we first  eliminate the noise by adding several random seeds $\{_s\boldsymbol{\mathcal{\xi}}^q_1, ..., _s\boldsymbol{\mathcal{\xi}}^q_k\}$. Then we choose one by calculating the maximum inlier matching number as $\boldsymbol{\mathcal{\xi}}^q_{t_1}$. The virtual pose experiences a \textit{Render-and-Compare} update process towards GT target $^*\boldsymbol{\mathcal{\xi}}^q$, varies from intermediate $\boldsymbol{\mathcal{\xi}}^q_{t_i}$ at step $i$ to final $\boldsymbol{\mathcal{\xi}}^q_{t_h}$ at step $h$. (b,c,d). The feature correspondences are visualized between the query and rendered image during the iterative refinement, where warmer colors indicate higher confidence. The matching results improve a lot along with the sequential adjustments.}
   \label{fig:main}
\end{figure}

An overview of the proposed method is exhibited in Figure~\ref{fig:main}. Given the textured 3D reconstruction built from aerial photography imaging, our objective is to estimate the 6-DoF poses $\{\boldsymbol{\mathcal{\xi}}^q\}$ for the altered perspective query images $\{ \mathbf{I}^q \}$ based on their noisy sensor priors $\{_s\boldsymbol{\mathcal{\xi}}^q\}$. To achieve this goal, we propose a novel three-stage \textit{Render-and-Compare} framework, which first generates prior poses (Section~\ref{subsec:initial}), then renders synthesis views and depthmaps for each pose hypothesis (Section~\ref{subsec:synthesis}), and finally conducts virtual pose correction until the final pose rendering aligns with the query image (Section~\ref{subsec:correction}).        

\subsection{Prior Pose Generation}
\label{subsec:initial}
The sensor pose acquisition for mobile phones can be divided into two parts: 1) For translation, the latitude and longitude coordinates in GPS are used as x-y values. Due to the low accuracy of GPS in the altitude direction, the x-y position is projected vertically onto a pre-built 3D surface and its floor height is recorded. The z-value is obtained by adding $1.5$ meters to the floor height to account for the hand-held gesture typically used when taking phone photos. 2) For rotation, we build $SO(3)$ matrix utilizing gravity and compass direction from mobile sensor.

Considering the sensor noises on the ground, we augment pose perturbations for smartphones. Specifically, we extend $\pm 5$ meters for x-y coordinate and supplement $\pm 60$ degrees for Euler angle yaw around the pose prior. We only adjust the yaw angle because the gravity sensor, which determines roll and pitch angles, shows high accuracy. The candidates of initial pose increase from $_s\boldsymbol{\mathcal{\xi}}^q$ to $\{_s\boldsymbol{\mathcal{\xi}}^q_1, ..., _s\boldsymbol{\mathcal{\xi}}^q_k\}$, where $k$ is the augment number. 


\subsection{View Synthesis}
\label{subsec:synthesis}
\begin{figure}[t]
   \centering
   \includegraphics[width=0.8\linewidth]{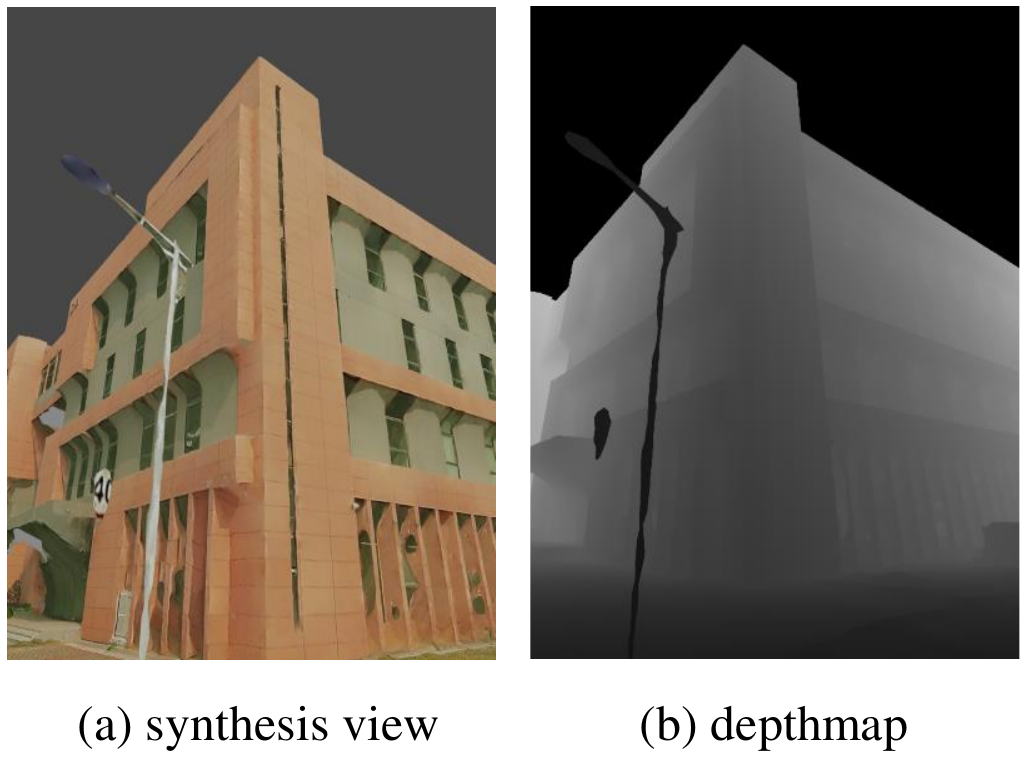}
   \caption{\textbf{Rendering result visualization.} An example of the rendering results is provided. (a) shows the synthesized view, while (b) illustrates the depthmap.}
   \label{fig:rendering}
\end{figure}

For a query $\mathbf{I}^q$, assuming that we have obtained all possible initials $\{_s\boldsymbol{\mathcal{\xi}}^q_1, ..., _s\boldsymbol{\mathcal{\xi}}^q_k\}$, we aim to acquire the corresponding textured renderings and depthmaps. Taking the time overhead into account, we apply Workbench Engine in Blender~\footnote{https://www.blender.org} with flat color mechanisms for fast rendering. Note that although we now use Blender to verify the \textit{Render-and-Compare} framework, the pipeline can be replaced with a more advanced rendering technique, such as Neural Radience Field (NeRF)~\cite{turki2022mega}. We denote the rendered image and depth for each virtual pose $_s\boldsymbol{\mathcal{\xi}}^q_i$ as $(_s\mathbf{I}^q_i, _s\mathbf{d}^q_i)$. An example of the rendering result, including synthesis view and depthmap, is shown in Figure~\ref{fig:rendering}.

\subsection{Pose Correction}
\label{subsec:correction}
The pose correction stage first selects the maximum likelihood pose $\boldsymbol{\mathcal{\xi}}^q_{t_1}$ from initial candidates $\{_s\boldsymbol{\mathcal{\xi}}^q_1, ..., _s\boldsymbol{\mathcal{\xi}}^q_k\}$. In detail, we employ state-of-the-art learned local features~\cite{detone2018superpoint} and matching strategies~\cite{sarlin2020superglue, sun2021loftr} to find 2D-2D correspondences between query $\mathbf{I}^q$ and renderings $\{ _s\mathbf{I}^q_i | i=1,...,k \}$, where $k$ is the seed number. A fundamental matrix is employed to prune possible incorrect correspondences. The candidate with the largest matching number wins the selection, denoted as:
\begin{equation}
	\label{equ:max-matching}
	_s\mathbf{I}^q \leftarrow \argmax \limits_{_s\mathbf{I}^q_i} \mathcal{M}_{2D-2D}^{num} (\mathbf{I}^q,_s\mathbf{I}^q_i),
\end{equation}
where $\mathcal{M}_{2D-2D}^{num}$ refers to the number of 2D-2D correspondences, $_s\mathbf{I}^q$ represents the synthesized image with the maximum number of matches selected from a set of synthesized images ${ _s\mathbf{I}^q_i }$, which participates in subsequent pose optimization iterations.

The $(\boldsymbol{\mathcal{\xi}}^q_{t_0}, \mathbf{I}^q_{t_0}, \mathbf{d}^q_{t_0})$ is regarded as the first ($t_0$) iteration during pose correction. The 2D-2D matches are then lift to 2D-3D using current pose estimation $\boldsymbol{\mathcal{\xi}}^q_{t_0}$ and the depth map $\mathbf{d}^q_{t_0}$. The next pose estimation $\boldsymbol{\mathcal{\xi}}^q_{t_1}$ is recovered by a PnP solver~\cite{haralick1994review} within LO-RANSAC~\cite{chum2003locally} loops. By repeating the iterative camera optimizations by $h$ times, the computed pose finally converges to an ultimate result $\boldsymbol{\mathcal{\xi}}^q = \boldsymbol{\mathcal{\xi}}^q_{t_h}$, as described in: 
\begin{equation}
	\label{equ:pipeline}
	\boldsymbol{\mathcal{\xi}}^q_{t_0} \xrightarrow{R\&C} \boldsymbol{\mathcal{\xi}}^q_{t_1}\xrightarrow{R\&C} \cdots \xrightarrow{R\&C} \boldsymbol{\mathcal{\xi}}^q_{t_h},
\end{equation}
where R\&C indicates the abbreviation of \textit{Render-and-Compare}.

Taking efficiency and effectiveness into consideration, we set $k=15, h=3$ in the experiment.

\section{Dataset}
\label{sec:dataset}
The released dataset \textit{AirLoc} includes a large urban area (approximately $100,000 m^2$), containing buildings, streets, and vegetation. In total, there are $1,970$ reference images and $1,432$ query images. The query images are captured from various viewpoints, including smartphones on the ground and UAVs in the near air. Moreover, the query images incorporate Day-and-Night conditions. Please see Figure~\ref{fig:dataset} for a visualization of the dataset.
 
\subsection{Reference Map Collection}
We capture high-resolution aerial image sequences with a five-eye oblique camera on a flight platform, using SHARE PSDK 102S\footnote{https://www.shareuavtec.com/ProductDetail/6519312.html} and DJI M300 RTK\footnote{https://www.dji.com/cn/matrice-300}, respectively. In order to fully and evenly cover the survey area, all flight paths are pre-planned in a grid fashion by the flight control system of DJI M300 RTK. Modern 3D reconstruction techniques are applied to build a textured mesh model and align it with the real geographic world with built-in RTK measurements. Note that, the camera has the ability to take both oblique and nadir photographs, ensuring that vertical surfaces are captured appropriately. Please refer to the supplementary material for a more detailed introduction.

\subsection{Query Image Collection}
We conduct query image acquisition using a smartphone and a micro aerial drone near the ground, i.e., HUAWEI Mate30 and DJI Phantom 4\footnote{https://www.dji.com/cn/phantom-4-pro}. The query images include multiple sessions during daytime and night-time. For each session, the SensorLog Application\footnote{http://sensorlog.berndthomas.net} is used to record raw in-build data. Since our method studies robust visual localization under noisy priors, we do not ensure that the hardware is synchronized and carefully calibrated for all sensors. Further details are provided in the supplementary material.


\subsection{Query GT Generation}
\begin{figure}[t]
   \centering
   \includegraphics[width=0.85\linewidth]{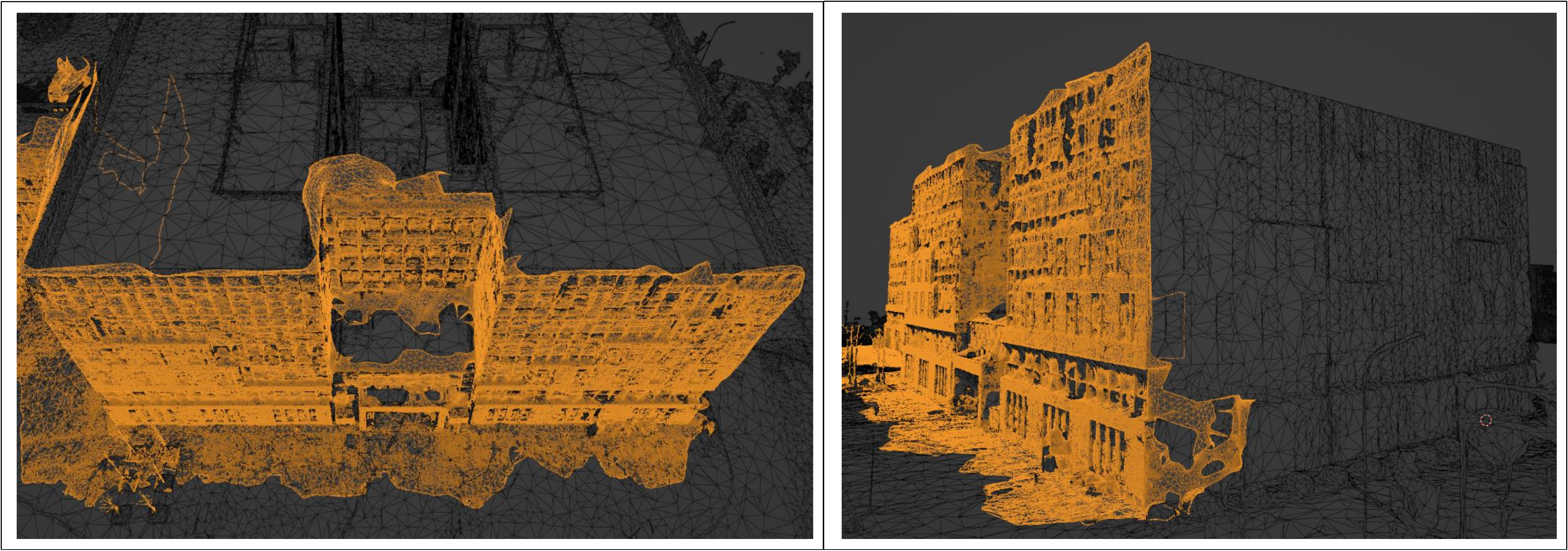}
   \caption{\textbf{Alignment quality of the aerial-to-ground reconstruction on \textit{AirLoc}.} The dark black model comes from aerial oblique photography, while the yellow model is built from a sequence of ground cellphone photos. The accuracy of the alignment can be observed in, for instance, the agreement of corners and edges. }
   \label{fig:regis_vis}
\end{figure}

\begin{figure}[t]
   \centering
   \includegraphics[width=0.95\linewidth]{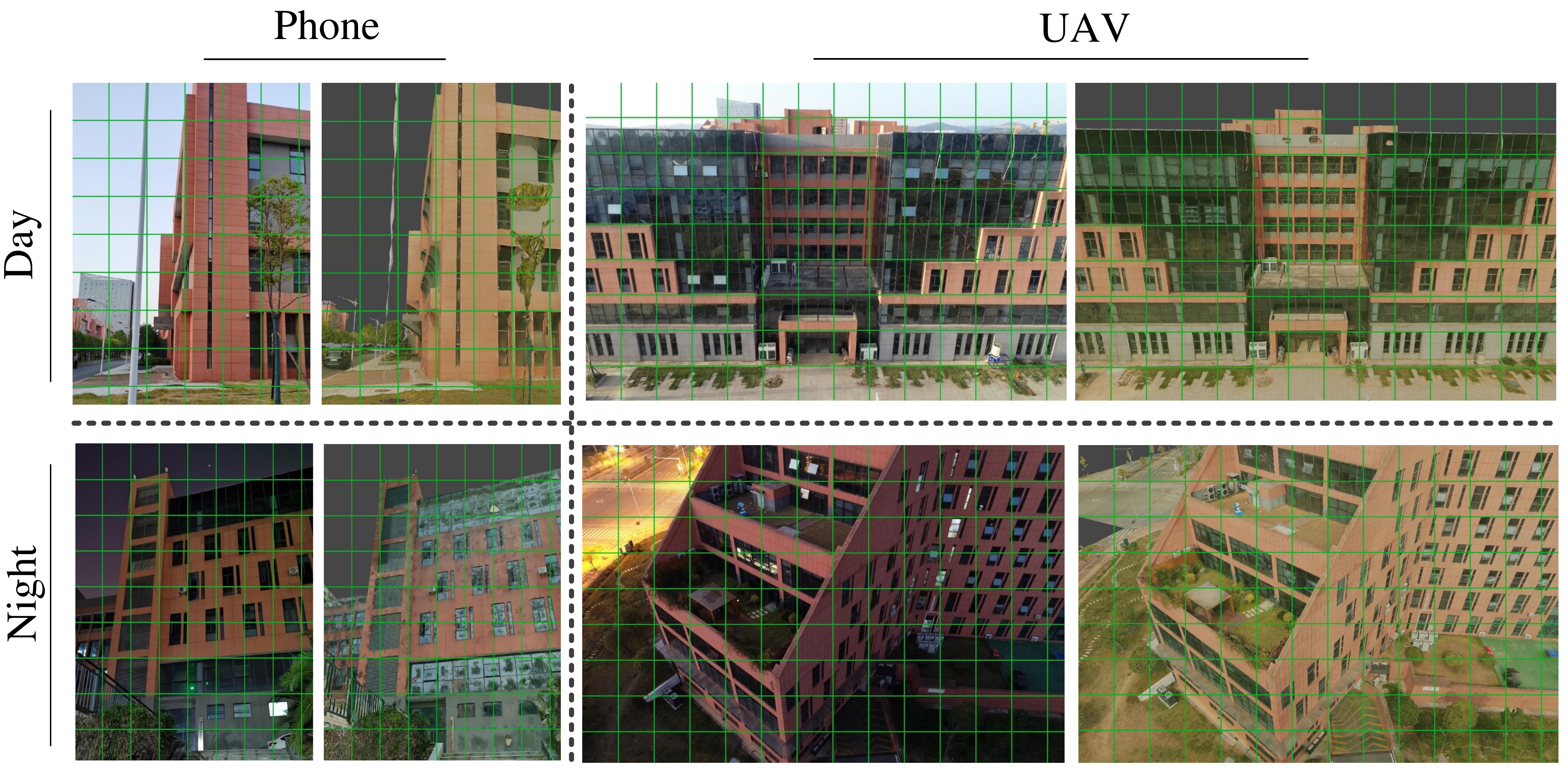}
   \caption{\textbf{GT poses quality on \textit{AirLoc}.} Pixel-aligned renderings of the estimated camera pose confirm that the poses are sufficiently accurate for our evaluation.}
   \label{fig:pixel_vis}
\end{figure}

We apply a scalable semi-automatic annotation to find the GT pose parameters $\{ \boldsymbol{\mathcal{\xi}}_i^q \}$ for query images. Our GT tool can label thousands of query poses, while only asking for dozens of manual assignments. In detail, Structure-from-Motion is first employed to reconstruct sparse point clouds for reference images and multiple query sequences. As ground and aerial images show remarkable visual differences, it is quite difficult to match keypoints even adopting state-of-the-art learning-based techniques~\cite{sarlin2020superglue, sun2021loftr}. We manually specify some iconic tie points across aerial-and-ground images, and conduct bundle adjustment for all individual SfM blocks. Finally, an integral 3D registration model is obtained, which owns camera poses for reference and query images. The accuracy of the GT poses is evaluated by median reprojection errors, which are $0.52$ pixels for the whole 3D model and $0.38$ for tie points, respectively. For visual inspection, the aerial and ground model registration quality is demonstrated in Figure~\ref{fig:regis_vis}. Besides, we randomly render images at the estimated GT poses using the textured mesh in Figure~\ref{fig:pixel_vis}. They appear in pixel-level alignment, supporting that the poses are accurate.  

\begin{table}[t]
\caption{\textbf{Visual localization results.} We report the recall at (25cm,2deg) / (50cm,5deg) / (1m,10deg). Our method is compared with HLoc~\cite{sarlin2019coarse} and MeshLoc~\cite{panek2022meshloc} with different feature-matching methods.}
\centering
\resizebox{0.48\textwidth}{!}{
\setlength\tabcolsep{2pt} 
    \begin{tabular}{l||ll} 
        \Xhline{3\arrayrulewidth}
        \multirow{2}{*}{Method} & \multicolumn{1}{c}{Day} & \multicolumn{1}{c}{Night} \\ 
        \cline{2-3} & \multicolumn{2}{c}{$(25cm, 2^{\circ})$ / $(50cm, 5^{\circ})$ / $(1m, 10^{\circ})$}   \\ 
        \Xhline{3\arrayrulewidth}
        \multicolumn{3}{c}{UAV} \\
        \hline
        HLoc~\textit{(SIFT + NN)} & 19.8 / 31.5 / 37.0 & \quad 5.2 / 62.6 / 80.0
                                                             \\
        HLoc~\textit{(SPP + NN)} & 41.6 / 52.9 / 56.3 & \quad 0.0 / 29.6 / 87.8 
                                                              \\
        HLoc~\textit{(SPP + SPG)} & 67.2 / 84.9 / 87.4 & \quad 0.0 / 53.9 / 100.0 
                                                                     \\
        \hline
        MeshLoc~\textit{(SPP + NN)} & 14.7 / 31.5 / 46.6 & \quad 1.7 / 29.6 / 60.9 
                                                         \\
        MeshLoc~\textit{(Patch2Pix)} & 19.3 / 34.5 / 48.7 & \quad 1.7 / 23.5 / 58.3 
                                                            \\    
        MeshLoc~\textit{(SPP + SPG)} & 72.7 / 80.0 / 82.8 & \quad 0.0 / 69.6 / 98.3 
                                                        \\
        MeshLoc~\textit{(LoFTR)} & 74.0 / 83.6 / 84.5 & \quad 0.0 / 40.9 / 90.4                                                                                    \\       \hline
        Ours~\textit{(SPP + SPG)} & \textbf{90.8} / \textbf{99.6} / \textbf{100.0} & \quad \textbf{90.0} / \textbf{100.0} / \textbf{100.0} 
                                                    \\
        Ours~\textit{(LoFTR)}
                   & 84.5 / 99.6 / 100.0 & \quad 81.7 / 99.1 / \textbf{100.0} 
                                                    \\ 
        \Xhline{3\arrayrulewidth}
        \multicolumn{3}{c}{Phone} \\
        \hline
        HLoc~\textit{(SIFT + NN)} & 0.0 / 0.0 / 0.0 & \quad 0.0 / 0.0 / 0.0 
                                                             \\
        HLoc~\textit{(SPP + NN)} & 0.0 / 0.0 / 0.0 & \quad 0.0 / 0.0 / 0.0 
                                                            \\
        HLoc~\textit{(SPP + SPG)} & 19.1 / 25.3 / 28.3 & \quad 19.1 / 25.1 / 27.8 
                                                            \\
        \hline
        MeshLoc~\textit{(SPP + NN)} & 0.0 / 0.0 / 0.0 & \quad 0.0 / 0.0 / 0.2 
                                                         \\
        MeshLoc~\textit{(Patch2Pix)} & 0.0 / 0.0 / 0.0 & \quad 0.0 / 0.0 / 0.6 
                                                            \\ 
        MeshLoc~\textit{(SPP + SPG)} & 14.5 / 22.8 / 26.9 & \quad 10.7 / 12.9 / 15.9 
                                                             \\
        MeshLoc~\textit{(LoFTR)} & 12.1 / 18.3 / 22.2 & \quad 11.1 / 15.3 / 18.6                                                      \\             
        \hline
        Ours~\textit{(SPP + SPG)} & \textbf{47.8} / 81.0 / 86.1 & \quad 24.5 / 56.2 / 68.6
                                                           \\
        Ours~\textit{(LoFTR)}   & 46.5 / \textbf{81.6} / \textbf{88.9} & \quad \textbf{28.2} / \textbf{60.7} / \textbf{74.4} 
                                                           \\
        \Xhline{3\arrayrulewidth}
    \end{tabular}
}
\label{tab:localization}
\end{table}

\section{Experiment}
\label{sec:experiment}
In this section, we introduce the selection of baselines and evaluation metrics on \textit{AirLoc} dataset in Section~\ref{subsec:baselines_metrics}. Experimental results and ablation studies are reported in Section~\ref{subsec:evaluation_results} and Section~\ref{subsec:ablation_studies}, respectively.

\subsection{Baselines and Metrics} 
\label{subsec:baselines_metrics}
\paragraph{Baselines}
Our approach is compared with state-of-the-art visual localization baselines, mainly HLoc~\cite{sarlin2019coarse, sarlin2020superglue} and MeshLoc~\cite{panek2022meshloc}, with different feature extractors and matchers, including sparse keypoint-based methods (SIFT~\cite{lowe2004distinctive}, SuperPoint(SPP)~\cite{detone2018superpoint}, Nearest Neighbors(NN), SuperGlue(SPG)~\cite{sarlin2020superglue}) and semi-dense methods (LoFTR~\cite{sun2021loftr}, Patch2Pix~\cite{zhou2021patch2pix}). For our \textit{Render-and-Compare} framework, SuperPoint(SPP)~\cite{detone2018superpoint}+SuperGlue(SPG)~\cite{sarlin2020superglue} and LoFTR~\cite{sun2021loftr} are adopted to match query images and virtual renderings. Note that all mentioned features are directly applied without fine-tuning or re-training on the \textit{AirLoc} dataset.

For a fair comparison, we do not employ global features to retrieve image pairs for HLoc and MeshLoc, as it is found that feature-based approaches (e.g., NetVLAD~\cite{arandjelovic2016netvlad}) suffered a lot in cross-view situations. Instead, we first obtain top-50 retrieval pairs by computing a modified Chamfer distance between prior observed points of the query and 3D observation of the reference. Then, the compass direction and GPS x-y coordinates are exploited to filter obvious wrong ranks. Supplemental materials provide more details.

\paragraph{Metrics}
We follow the standard localization evaluation procedure~\cite{sattler2018benchmarking}, and report the localization recall at thresholds $(25cm, 2^{\circ})$, $(50cm, 5^{\circ})$, and $(1m, 10^{\circ})$. The results are divided into four splits by Day-and-Night conditions and different capturing devices (namely, phone and UAV).

\subsection{Evaluation Results and Analysis}
\label{subsec:evaluation_results}
The localization results are reported in Table~\ref{tab:localization}. Our method substantially outperforms HLoc~\cite{sarlin2019coarse, sarlin2020superglue} and MeshLoc~\cite{panek2022meshloc} by a large margin under all conditions, even employing the same local features (e.g., SuperPoint(SPP)~\cite{detone2018superpoint}+SuperGlue(SPG)~\cite{sarlin2020superglue} and LoFTR~\cite{sun2021loftr}). For instance, in daytime visual localization of UAVs using SPP+SPG for feature matching, the recall of the proposed method at $(25cm, 2^{\circ})$, $(50cm, 5^{\circ})$, and $(100cm, 10^{\circ})$ are 90.8\%, 99.6\%, and 100.0\%, respectively. In contrast, the second-ranked MeshLoc achieves only 72.7\%, 80.0\%, and 82.8\%. The advantage of our method is more evident for nighttime visual localization of UAVs. While other baseline methods essentially fail to localize under the strict $(25cm, 2^{\circ})$ criterion, the proposed method still achieves a localization recall of 90.0\%.

\begin{figure}[t]
    \centering
    \includegraphics[width=1.0\linewidth]{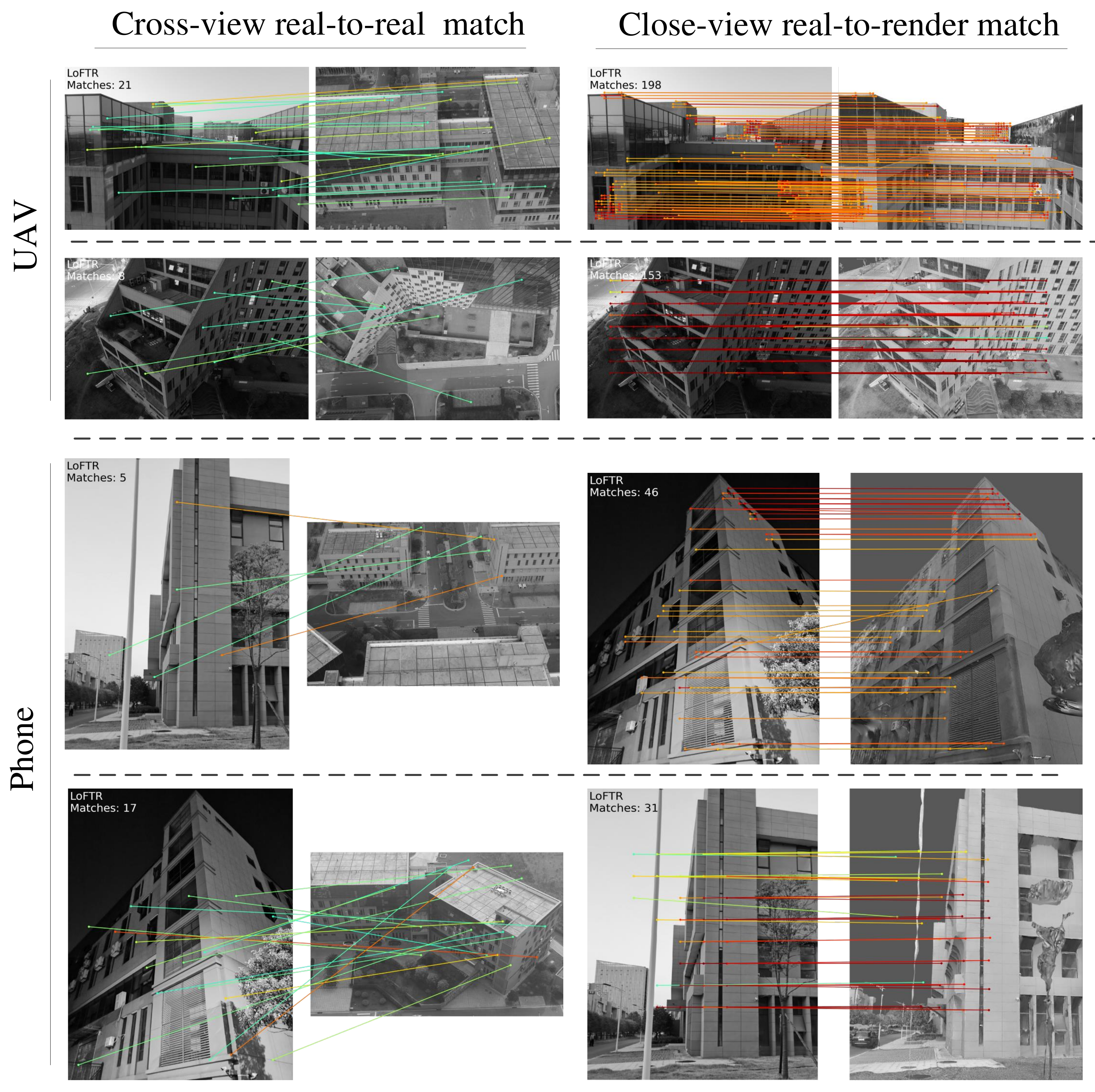}
    
    \caption{\textbf{Mechanism analysis of advantage of our method.} The left column shows that state-of-the-art learned feature LoFTR~\cite{sun2021loftr} cannot establish correspondences for real image pairs sharing notable viewpoint differences. The right column illuminates that LoFTR~\cite{sun2021loftr} is good at matching real-and-synthetic views with similar poses. Warmer colors show higher confidence.}
    \label{fig:matches}
\end{figure}

The results fully demonstrate the capability of our \textit{Render-and-Compare} design for camera localization. We attribute the success to the following mechanism: feature matching between real-to-render pairs with similar viewpoints owns overwhelming advantages over read-to-real pairs with distinct perspectives. To visually explain this phenomenon, the LoFTR~\cite{sun2021loftr} feature-matching results are provided in Figure~\ref{fig:matches}. It is obvious that local matching between viewpoint-closed synthetic-to-real views takes the lead over viewpoint-cross real-to-real views in terms of matching number and accuracy.

It is also worth noting that under daytime and nighttime conditions, the localization results of mobile phones using LoFTR are only 46.5\% and 28.2\% for $(25cm, 2^{\circ})$ criterion, respectively. This indicates that the problem of cross-view 6-DoF visual localizationhas not been completely solved and requires further research. Some typical failure cases are provided in the supplementary material.

\subsection{Ablation Studies}
\label{subsec:ablation_studies}
\paragraph{Iteration number}
As a sequential update method, we exploit the influence of the iteration number. Increasing the number increases the localization at all thresholds, as reported in Table~\ref{tab:ablation_iter}. Considering that well-matured OpenGL-based rendering technology, which is optimized for real-time performance on GPUs~\cite{panek2022meshloc}, this \textit{Render-and-Compare} solution has the potential to efficiently achieve better results with more refinements.

\begin{table}[t]
\caption{\textbf{Ablation studies} Different iteration number $\{ 1,2,3 \}$ in our \textit{Render-and-Compare} framework is compared with (25cm,2deg) / (50cm,5deg) / (1m,10deg) metrics.}
\centering
\resizebox{0.48\textwidth}{!}{
\setlength\tabcolsep{2pt} 
    \begin{tabular}{l||ll} 
        \Xhline{3\arrayrulewidth}
        \multirow{2}{*}{Method} & \multicolumn{1}{c}{Day} & \multicolumn{1}{c}{Night} \\ 
        \cline{2-3} & \multicolumn{2}{c}{$(25cm, 2^{\circ})$ / $(50cm, 5^{\circ})$ / $(1m, 10^{\circ})$}   \\ 
        \Xhline{3\arrayrulewidth}
        \multicolumn{3}{c}{UAV} \\
        \hline
        LoFTR~\textit{(iter 1)} & 73.1 / 98.3 / 100.0 & \quad 47.8 / 81.7 / 96.5
                                                             \\
        LoFTR~\textit{(iter 2)} & 75.2 / 99.4 / 100.0 & \quad 79.1 / 96.5 / 99.1 
                                                            \\
        LoFTR~\textit{(iter 3)} & 84.5 / 99.6 / 100.0 & \quad 81.7 / 99.1 / 100.0                                                                  \\
        \hline
        SPP+SPG~\textit{(iter 1)} & 82.3 / 99.2 / 100.0 & \quad 89.6 / 100.0 / 100.0     
                                                         \\
        SPP+SPG~\textit{(iter 2)} & 84.9 / 99.6 / 100.0 & \quad 89.7 / 100.0 / 100.0     
                                                              \\
        SPP+SPG~\textit{(iter 3)} & 90.8 / 99.6 / 100.0 & \quad 90.0 / 100.0 / 100.0     
                                                             \\   
        \Xhline{3\arrayrulewidth}
        \multicolumn{3}{c}{Phone} \\
        \hline
        LoFTR~\textit{(iter 1)} & 12.6 / 37.7 / 68.3 & \quad 5.3 / 20.5 / 46.4
                                                             \\
        LoFTR~\textit{(iter 2)} & 37.3 / 75.4 / 86.9 & \quad 17.6 / 50.1 / 65.5
                                                            \\
        LoFTR~\textit{(iter 3)} & 46.5 / 81.6 / 88.9 & \quad 28.2 / 60.7 / 74.4                                                                 \\
        \hline
        SPP+SPG~\textit{(iter 1)} & 12.0 / 36.0 / 67.4 & \quad 3.8 / 14.1 / 39.3    
                                                         \\
        SPP+SPG~\textit{(iter 2)} & 37.0 / 73.6 / 84.1 & \quad 15.2 / 43.1 / 58.2     
                                                              \\
        SPP+SPG~\textit{(iter 3)} & 47.8 / 81.0 / 86.1 & \quad 24.5 / 56.2 / 68.6     
                                                        \\  
        \Xhline{3\arrayrulewidth}
    \end{tabular}
}
\label{tab:ablation_iter}
\end{table}

\begin{table}[t]
\caption{\textbf{Ablation studies.} Different seed augmentation strategies are compared with (25cm,2deg) / (50cm,5deg) / (1m,10deg) metrics using LoFTR~\cite{sun2021loftr}.}
\centering
\resizebox{0.48\textwidth}{!}{
\setlength\tabcolsep{2pt} 
    \begin{tabular}{l||ll} 
        \Xhline{3\arrayrulewidth}
        \multirow{2}{*}{Initial augmentation} & \multicolumn{1}{c}{Day} & \multicolumn{1}{c}{Night} \\ 
        \cline{2-3} & \multicolumn{2}{c}{$(25cm, 2^{\circ})$ / $(50cm, 5^{\circ})$ / $(1m, 10^{\circ})$}   \\ 
        \Xhline{3\arrayrulewidth}
        \multicolumn{3}{c}{UAV} \\
        \hline
        Seed~\textit{(not used)} & 81.9 / 98.3 / 100.0 & \quad 79.1 / 96.5 / 99.1 
                                                             \\
        Seed~\textit{(translation x-y)} & 81.9 / 99.6 / 100.0 & \quad 79.1 / 96.5 / 100.0 
                                                             \\
        Seed~\textit{(rotation yaw)} & 81.9 / 99.2 / 100.0 & \quad 81.7 / 99.1 / 100.0 
                                                            \\
        Seed~\textit{(both)} & 84.5 / 99.6 / 100.0 & \quad 81.7 / 99.1 / 100.0 
                                                          \\
        \Xhline{3\arrayrulewidth}
        \multicolumn{3}{c}{Phone} \\
        \hline
        Seed~\textit{(not used)} & 5.8 / 16.3 / 35.1 & \quad 9.1 / 28.5 / 38.6 
                                                         \\
        Seed~\textit{(translation x-y)} & 36.8 / 46.5 / 68.5 & \quad 18.3 / 46.8 / 58.3 
                                                         \\
        Seed~\textit{(rotation yaw)} & 34.1 / 65.3 / 72.6 & \quad 21.3 / 44.3 / 60.0 
                                                              \\
        Seed~\textit{(both)} & 46.5 / 81.6 / 88.9 & \quad 28.2 / 60.7 / 74.4 
                                                             \\
        \Xhline{3\arrayrulewidth}
    \end{tabular}
}
\label{tab:ablation_seed}
\end{table}

\paragraph{Seed initialization.}
We evaluate the effect with or without translation and rotation augmentations of the initialization. As shown in Table~\ref{tab:ablation_seed}, both parameters significantly influence the final localization accuracy, especially for the phone data split, which proves the effectiveness of the proposed initial pose enhancement strategy.



\section{Conclusion}
\label{sec:conclusion}


In this paper, we present a new approach for cross-view 6-DoF localization from noisy priors. To overcome the difficulties of image matching over distinctive viewpoints, we propose a \textit{Render-and-Compare} framework to sequentially refine the pose estimation. The localization performance is evaluated on a new aerial-to-ground dataset with an oblique photography map. Our results demonstrate significant improvements compared to state-of-the-art localization methods.

\textbf{Acknowledgements.} 
The authors would like to acknowledge the support from the Natural Science Foundation of Hunan Province of China (2020JJ5671) and the National Natural Science Foundation of China (No. 62171451, No. 62101576, No. 62071478).


\bibliographystyle{IEEEtran} 
\bibliography{IEEEabrv, icme2023CameraReady}

\clearpage

\appendix
\section*{Supplementary Material}



\begin{figure}[ht]
   \centering
   \includegraphics[width=0.85\linewidth]{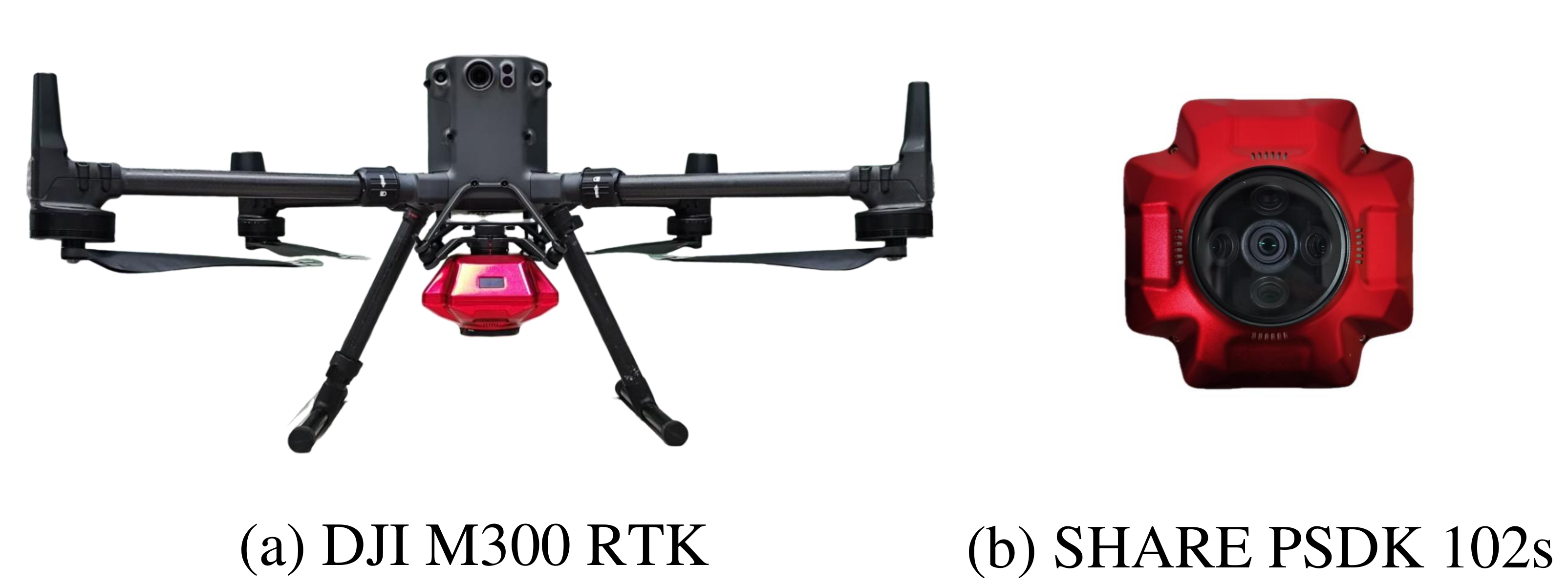}
   \caption{\textbf{Reference-capturing devices.} (a). The UAV platform DJI M300 RTK. (b). The lens details of the equipped SHARE PSDK 102s.}
   \label{fig:capture_device}
\end{figure}

\begin{figure}[ht]
   \centering
   \includegraphics[width=0.85\linewidth]{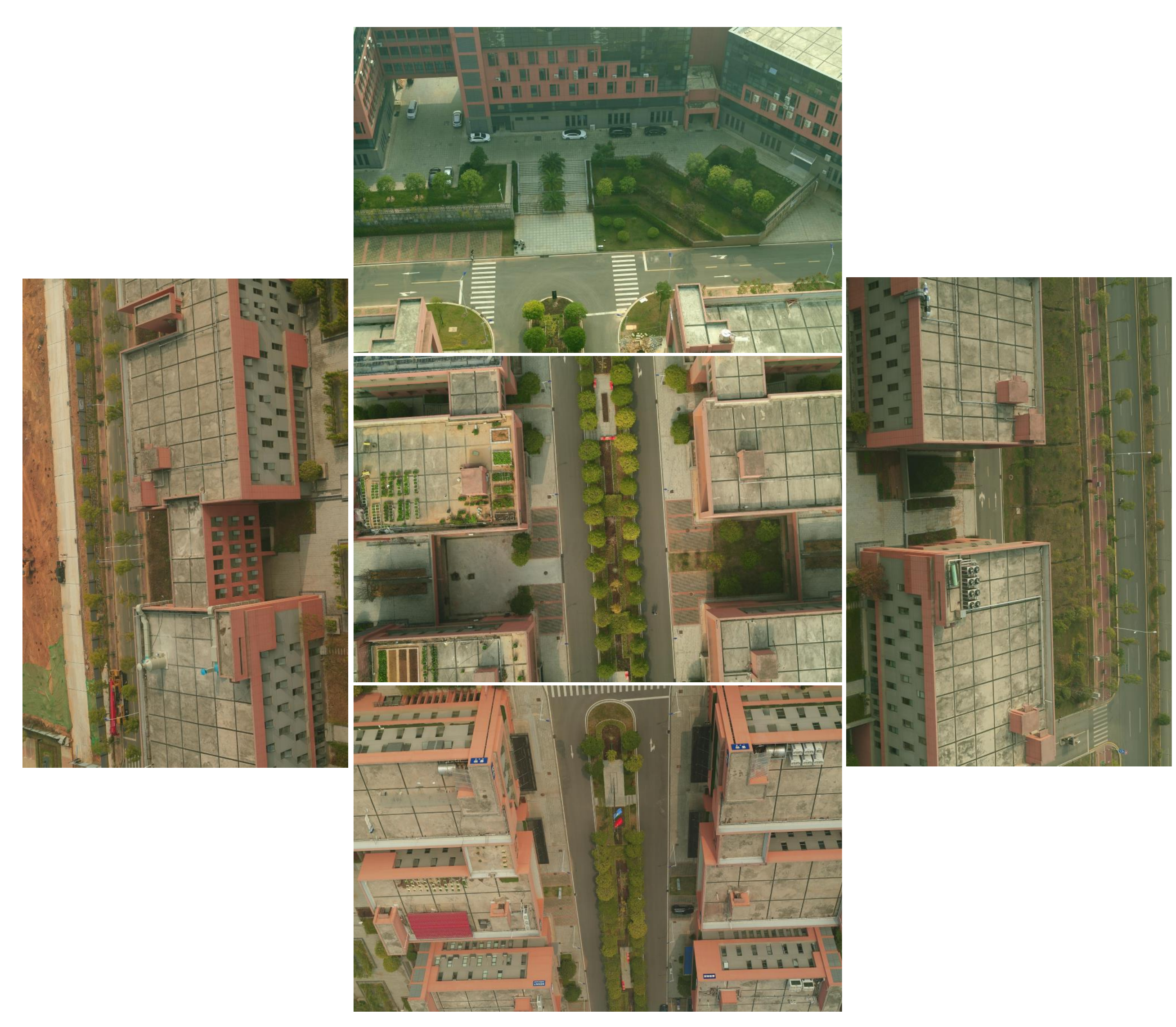}
   \caption{\textbf{Images captured by the SHARE 102s at one shot.} The five-eye camera owns forward, backward, left, right, and downward directions.}
   \label{fig:five_eye}
\end{figure}

\begin{figure}[t]
   \centering
   \includegraphics[width=0.85\linewidth]{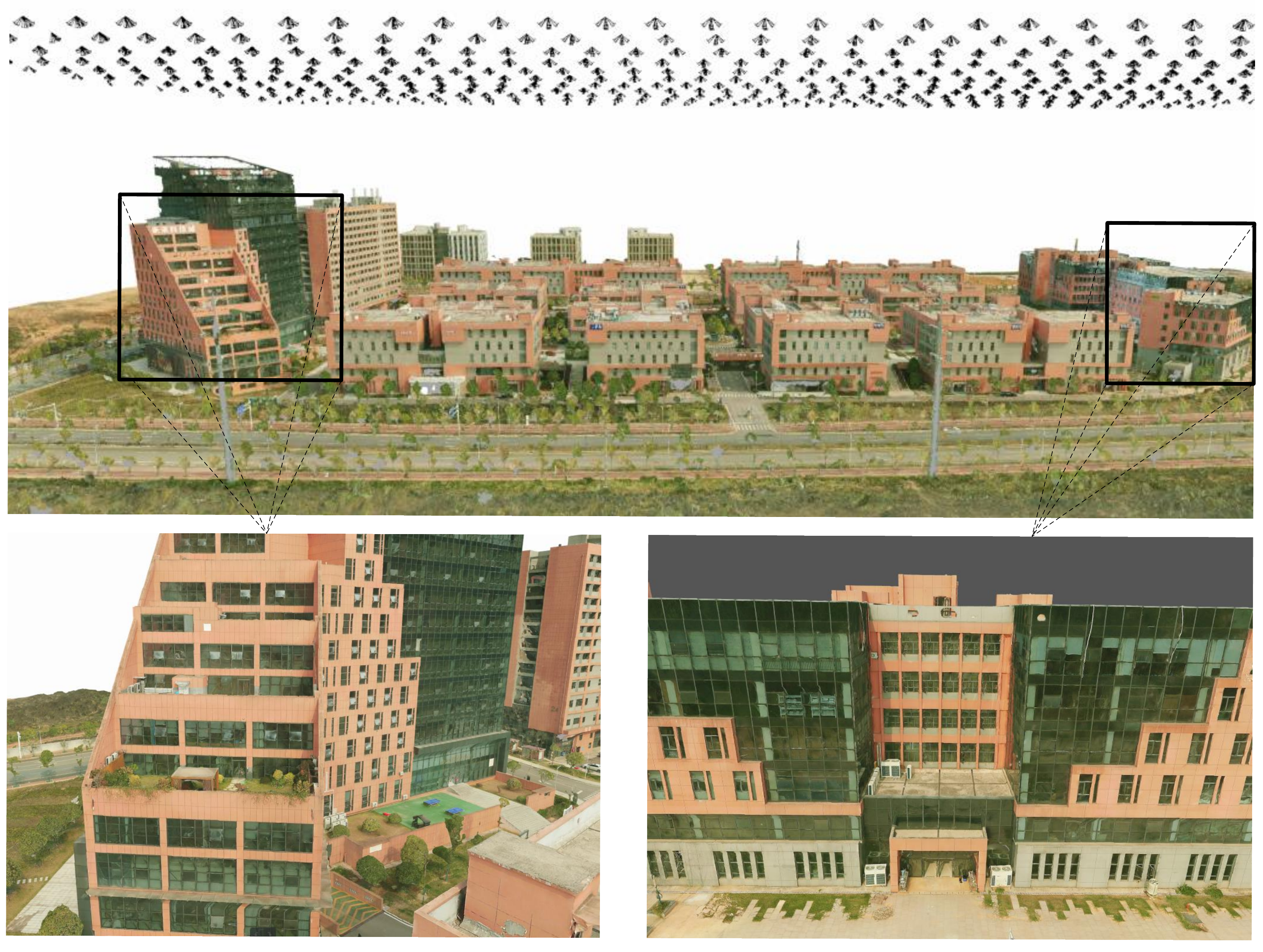}
   \caption{\textbf{Reference map reconstruction.} Given a large number of aerial oblique photographs, we produce a high-quality 3D surface reconstruction. Please zoom in to see the details.}
   \label{fig:reference_map}
\end{figure}

\begin{figure}[t]
   \centering
   \includegraphics[width=0.85\linewidth]{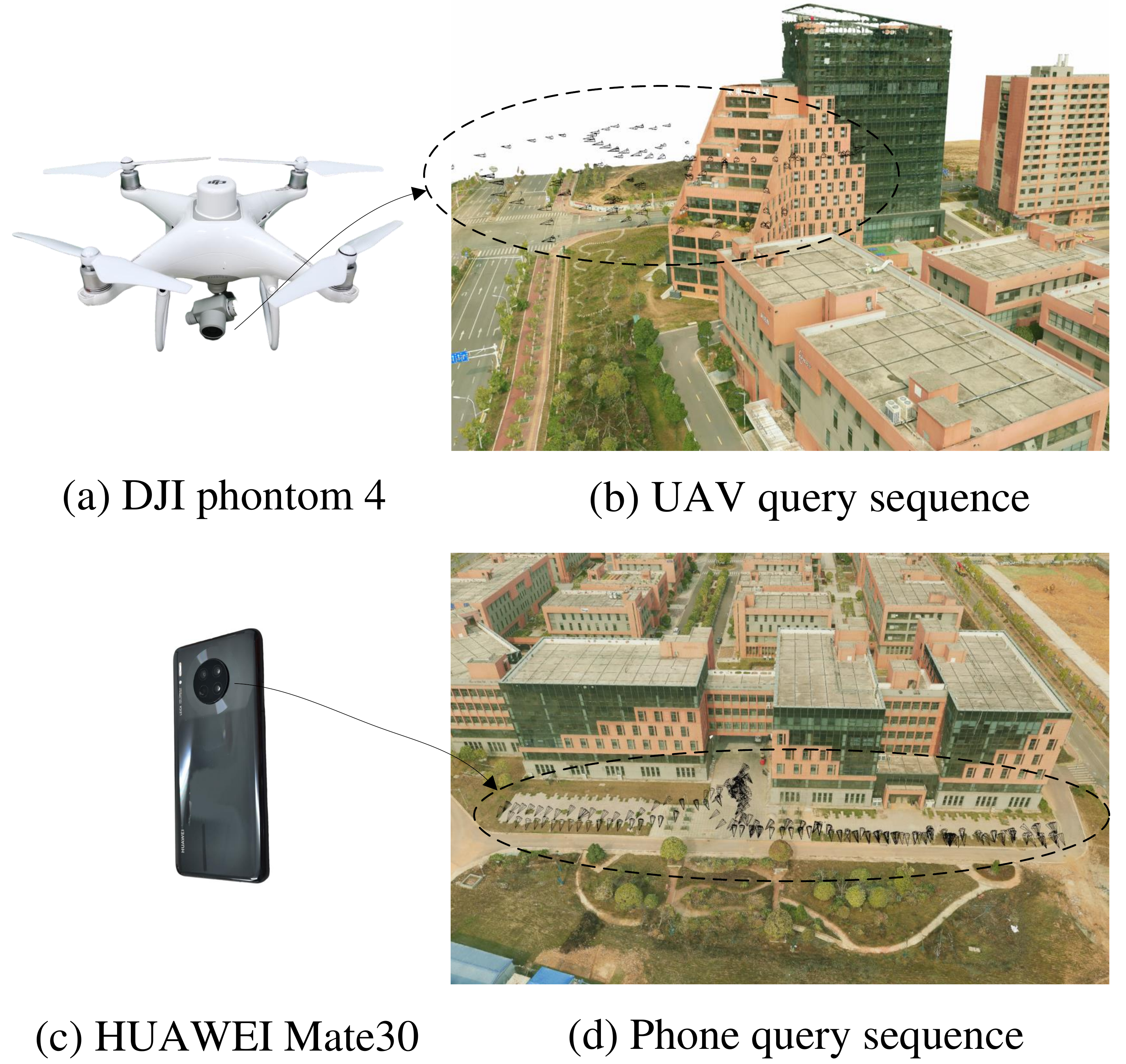}
   \caption{\textbf{Query image generation.} (a,c). The capture devices include a drone and a cellphone. (d,b). The black quadrangular pyramids inside the dotted line demonstrate the capture positions and orientations.}
   \label{fig:query}
\end{figure}

\begin{figure}[t]
   \centering
   \includegraphics[width=0.85\linewidth]{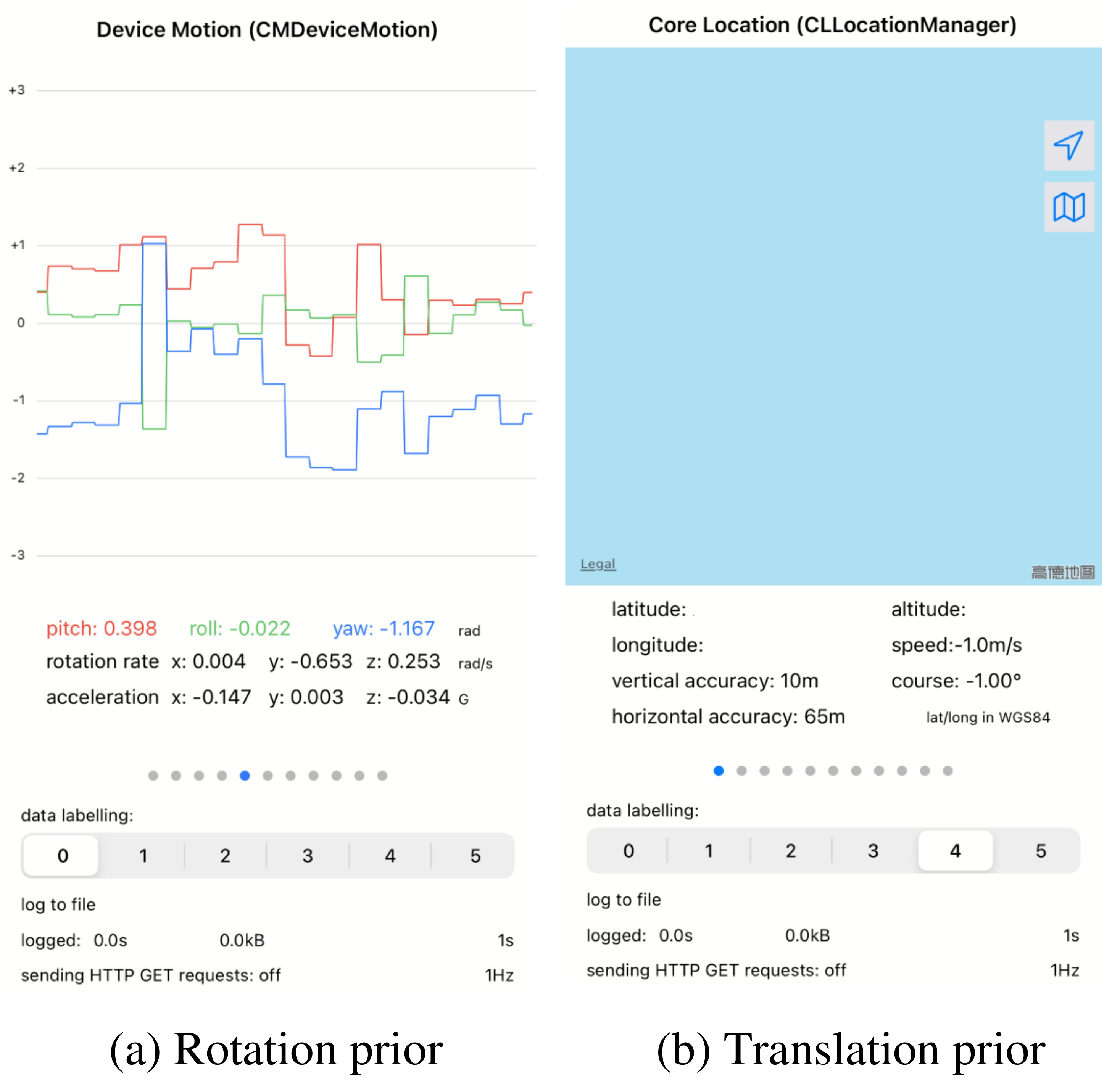}
   \caption{\textbf{Sensorlog interfaces for recording rotation and translation prior.} (a). The Euler angles change along the movement of the mobile phone. The pitch, roll, and yaw components are drawn in red, green and blue, respectively. The angular velocity and acceleration are also reported for all timestamps. (b). The GPS information, including latitude, longitude, and altitude, is noted in detail. For anonymous reasons, we remove the display of the specific location information. The APP also provides measurement accuracy in terms of vertical and horizontal.}
   \label{fig:sensorlog}
\end{figure}

\section{Dataset Details}
\label{sec:dataset}

\subsection{Reference Map Collection}
We collect the reference images via aerial oblique photography, the capture devices are shown in Figure~\ref{fig:capture_device}. Specifically, we employ a quadcopter drone, DJI M300 RTK\footnote{https://www.dji.com/cn/matrice-300}, which is equipped with a cutting-edge oblique five-eye camera SHARE PSDK 102s\footnote{https://www.shareuavtec.com/ProductDetail/6519312.html}, to stably capture high-resolution aerial image sequences. An example of a group of five-eye images is illuminated in Figure~\ref{fig:five_eye}. Based on the data collection, we apply modern 3D reconstruction techniques to recover the textured 3D reference model, which is visualized in Figure~\ref{fig:reference_map}.

\subsection{Query Image Collection}
We capture the query photos using a small drone near the ground and a smartphone on the ground, as shown in Figure~\ref{fig:query}. As for prior information, DJI drones directly save the sensor data in the EXIF file. In order to provide pose prior for cellphones, we record the raw in-build sensor data via SensorLog Application\footnote{http://sensorlog.berndthomas.net}. We pick an interface visualization from a specific time and place of SensorLog, as reported in Figure~\ref{fig:sensorlog}.

\begin{figure}[t]
   \centering
   \includegraphics[width=0.85\linewidth]{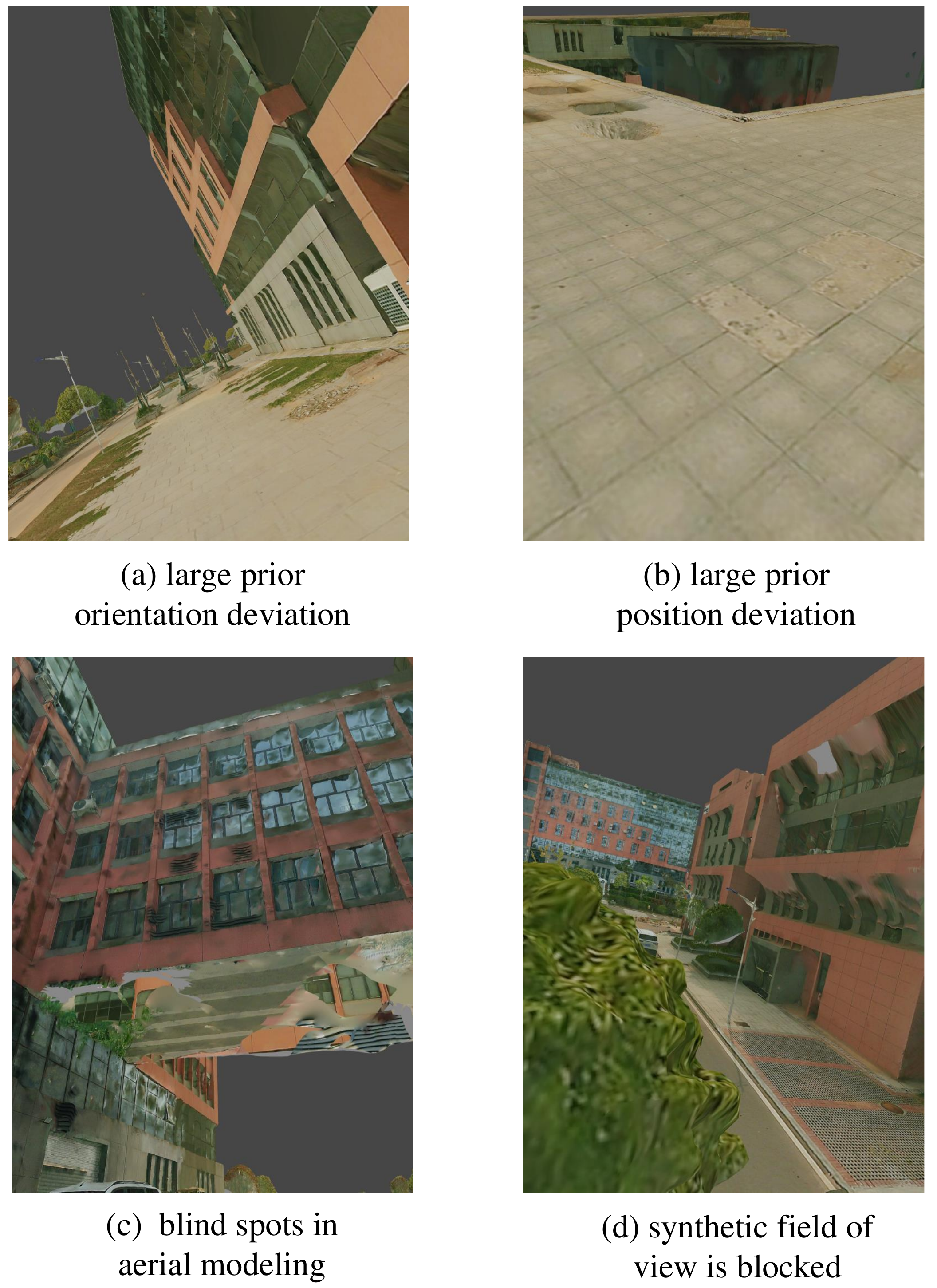}
   \caption{\textbf{Failure Cases of our method.} (a). The prior orientation deviation is too large. (b). The prior position deviation is too large. (c). there are blind spots in aerial modeling. (d). The field of view of the synthetic image is blocked.}
   \label{fig:sensorlog}
\end{figure}

\section{Experiment Details}
\label{sec:experiments}

\subsection{Prior Guided Image Retrieval}
Given a query image $\mathbf{I}^q$, HLoc~\cite{sarlin2019coarse, sarlin2020superglue} and MeshLoc~\cite{panek2022meshloc} need to find a covisible set $S_{covis}= \{ \mathbf{I}^r_1,\mathbf{I}^r_2,...,\mathbf{I}^r_l \}$ against reference collections $\{ \mathbf{I}^r \}$, where $l$ is the the number of retrieved items. One typical pipeline is to first map images $\{ \mathbf{I}^q, \mathbf{I}^r \}$ into a certain compact feature space via an embedding function $\boldsymbol{f}(\cdot)$. Then nearest neighbors of $\mathbf{I}^q$ is searched by a defined similarity distance $\boldsymbol{d}(\boldsymbol{f}(\mathbf{I}^q),\boldsymbol{f}(\mathbf{I}^r))$.

However, these retrieval methods~\cite{arandjelovic2016netvlad, ge2020self} completely rely on the distinctiveness of the global features. We found that if images shared notable viewpoint changes, like in \textit{AirLoc} dataset, they were prone to failure. For a fair comparison, we implement a prior guided image retrieval algorithm for HLoc and MeshLoc.

Our algorithm is based on the following assumption: if two images observe the same structure, the 3D point clouds they observe should be close. To this end, we back-project depthmaps of query images $\{ \mathbf{I}^q \}$ via prior poses and reference images $\{ \mathbf{I}^r \}$ via registered poses. After uniformly random sampling to $2048$, we obtain the observation point as $\{ \mathbf{P}^q \}$ for queries and $\{ \mathbf{P}^r \}$ for references. Please note that, in order to improve the robustness of the sensor noise, we add pose perturbations around the sensor prior and concatenate back-projection points of all seeds.

We calculate the back-projection point distance by a modified Chamfer Distance $MCD$. The raw Chamfer Distance $CD$ is usually used in the calculation of the point distance, which is defined as
\begin{equation}
\label{eq:raw chamfer distance}
    CD(\mathbf{P}^q, \mathbf{P}^r) = \frac{1}{|\mathbf{P}^{q}|}\sum_{q \in \mathbf{P}^q}\min_{r \in \mathbf{P}^r}||q-r||^2_2.
\end{equation}

To remove the background points, instead of minimizing the sum over all query points, we only select the smallest $k$ values, where $k$ represents the expected number. Without loss of generality, assume that the terms of the distance in Equation~\ref{eq:raw chamfer distance} have been re-ordered in ascending order, the modified Chamfer Distance is defined as: 

\begin{equation}
    MCD(\mathbf{P}^{q*}, \mathbf{P}^r) = \frac{1}{|\mathbf{P}^{q*}|}\sum_{q \in \mathbf{P}^{q*}}\min_{r \in \mathbf{P}^r}||q-r||^2_2,
\end{equation}
where $P^{q*}$ denotes the prioritized 50 query points rearranged in ascending order.

Based on the calculation of the modified Chamfer Distance $MCD$, we can obtain $top$-50 nearest neighbors $\{ \mathbf{I}^r_1,\mathbf{I}^r_2,...,\mathbf{I}^r_{50} \}$ for each query image $\mathbf{I}^q$. In order to solve the problem of direction ambiguity, such as point observation upon front and back of the wall, we use yaw and x-y to further filter and sort. Specifically, if the reference image is $90$ degrees of yaw or $100$ meters of x-y away from the query prior, we remove these retrieval candidates.

\subsection{Failure Cases of our method}
Being dependent on iterative \textit{Render-and-Compare} procedure, our method inherently suffers from very bad sensor prior and blind spots of reconstruction and view occlusions of rendering.

{



}

\end{document}